\documentclass[10pt,twocolumn,letterpaper]{article}

\usepackage{cvpr}
\makeatletter
\@namedef{ver@everyshi.sty}{}
\makeatother
\usepackage{graphicx}
\usepackage{amsmath}
\usepackage{amssymb}
\usepackage{booktabs}
\usepackage{lipsum,adjustbox}
\usepackage{multirow}
\usepackage[normalem]{ulem}
\useunder{\uline}{\ul}{}
\usepackage{hyperref}
\hypersetup{breaklinks,colorlinks}

\usepackage[capitalize]{cleveref}
\crefname{section}{Sec.}{Secs.}
\Crefname{section}{Section}{Sections}
\Crefname{table}{Table}{Tables}
\crefname{table}{Tab.}{Tabs.}

\begin{document}

\title{Distilling Calibrated Student from an Uncalibrated Teacher}

\author{Ishan Mishra\\
Indian Institute of Technology, Jodhpur\\
Jodhpur, Rajasthan, India\\
{\tt\small mishra.10@iitj.ac.in}
\and
Sethu Vamsi Krishna\\
Indian Institute of Technology, Jodhpur\\
Jodhpur, Rajasthan, India\\
{\tt\small krishna.3@iitj.ac.in}
\and
Deepak Mishra\\
Indian Institute of Technology, Jodhpur\\
Jodhpur, Rajasthan, India\\
{\tt\small dmishra@iitj.ac.in}
}
\maketitle
\begin{abstract}

Knowledge distillation is a common technique for improving the performance of a shallow student network by transferring information from a teacher network, which in general, is comparatively large and deep. These teacher networks are pre-trained and often uncalibrated, as no calibration technique is applied to the teacher model while training. Calibration of a network measures the probability of correctness for any of its predictions, which is critical in high-risk domains. In this paper, we study how to obtain a calibrated student from an uncalibrated teacher. Our approach relies on the fusion of the data-augmentation techniques, including but not limited to cutout, mixup, and CutMix, with knowledge distillation. We extend our approach beyond traditional knowledge distillation and find it suitable for Relational Knowledge Distillation and Contrastive Representation Distillation as well. The novelty of the work is that it provides a framework to distill a calibrated student from an uncalibrated teacher model without compromising the accuracy of the distilled student. We perform extensive experiments to validate our approach on various datasets, including CIFAR-10, CIFAR-100, CINIC-10 and TinyImageNet, and obtained calibrated student models. We also observe robust performance of our approach while evaluating it on corrupted CIFAR-100C data.

\end{abstract}

\section{Introduction}
\label{sec:intro}

Recent advances in deep learning have made deep neural networks  (DNNs)~\cite{he2016deep, huang2017densely, tan2019efficientnet}
achieve human-like performance in terms of accuracy. However, these models are often overconfident ~\cite{guo2017calibration, minderer2021revisiting} and deploying them in high-risk domains like medical, astronomical, autonomous vehicles, etc., becomes unsafe. 
Thus, the calibration of DNNs becomes critical for trustworthy AI, which helps in aligning the model's confidence with its accuracy. Various techniques like vector scaling \cite{guo2017calibration,nixon2019measuring}, matrix scaling \cite{guo2017calibration}, temperature scaling \cite{ding2021local, nixon2019measuring, kumar2019verified}, label smoothing \cite{chen2020investigation, muller2019does}, mixup \cite{zhang2018mixup, thulasidasan2019mixup}, CutMix \cite{yun2019cutmix} etc. help in calibrating the DNNs. Moreover, techniques like cutout, mixup, and CutMix, categorised under data-augmentation techniques, have also shown to make the DNN robust. However, training deep architectures with these techniques incurs overhead, while shallow architectures cannot match the accuracy of their deeper counterparts, even with augmentation. Thus, a method to transfer the knowledge from deep architectures to a shallow architecture while also improving its calibration is desirable.

Knowledge distillation (KD) \cite{hinton2015distilling} is a technique that enhances the performance of a student network by incorporating the knowledge from a teacher network. Generally, the student network is shallow, and the teacher network is a deeper network pre-trained on a large-scale dataset like ImageNet. KD is used extensively in real-life scenarios where computational complexity is a critical factor. The teacher model being a deep architecture is often expensive to deploy on a local machine, while a shallow student architecture, trained from scratch, does not deliver the same performance. Thus, KD helps in increasing the accuracy of the shallow student network. Various modifications over traditional KD 
are reported in past to boost the performance of the student network, for example, Relational KD (RKD)~\cite{park2019relational}, Contrastive Representation Distillation (CRD)~\cite{tian2019crd}, etc. However, improving the accuracy of the student without paying attention to its confidence makes it uncalibrated. Stanton \etal\cite{stanton2021does} analyzed the accuracy and fidelity of the student in the view of KD. 
\cref{fig:intro-cali} shows the poor calibration of the student distilled using conventional KD. 

Some works \cite{muhammad2021mixacm, das2020empirical}
show the distillation of a well-calibrated student from an already-calibrated teacher. However, large models with millions of parameters, quite common these days, are often unclibrated and calibrating such pre-trained model is expensive due to the involvement of huge datasets. We, therefore, ask a question - \textit{can a calibrated student be distilled from an uncalibrated teacher?}, and try to find its answer through our approach.

Following are the main contributions and findings of our paper:
\begin{itemize}
    \item We develop a simple approach to distilled a well-calibrated student network from an uncalibrated teacher. We use scaled version of KL-divergence loss with the augmentation loss to improve the calibration of the student.
    \item We propose a generalised framework that can easily be integrated with a variety of distillation and augmentation techniques. We perform experiments using three different augmentation techniques namely mixup, CutMix, cutout and with different distillation techniques like RKD and CRD.
    \item We perform experiments on various datasets, including CIFAR-10, CIFAR-100, CINIC-10, and TinyImageNet and obtain a student which is even better than teacher in terms of calibration.
    \item We perform experiments on corrupted (CIFAR-100C) to demonstrate the robustness of the distilled student network using our approach. 
\end{itemize}
\section{Related Work}
\subsection{Knowledge Distillation}
The first instance of knowledge transfer was reported by Bucilua \etal \cite{bucilua2006model}, where a single student is distilled from an ensemble of networks. Ba and Caruana \cite{ba2014deep} used the knowledge of a deep neural network to increase the accuracy of a shallow network.
The term ``Knowledge Distillation'' was coined by Hinton \etal \cite{hinton2015distilling} with the idea of minimizing the KL loss between the softened probabilities from the last layer of the teacher and student model. Broadly, KD is classified into offline, online \cite{guo2020online, wu2021peer, zhang2018deep}, and self distillation \cite{xu2020knowledge, zhang2019your, mobahi2020self} based on the pre-trained teacher. Researchers \cite{romero2014fitnets, zagoruyko2016paying} have also explored minimizing the loss between the intermediate layers rather than the last layer, which finds the similarity between the feature representation of the teacher and student.
Yim \etal \cite{yim2017gift} proposed a method of distilling relational knowledge from the teacher by using Gram matrix between the feature maps of first and last layers of the teacher model. Park \etal \cite{park2019relational} proposed RKD aiming to transfer knowledge by finding the relation between different instances. They introduced two new losses, distance-wise and angle-wise loss, to minimize the structural difference in relations. Beyer \etal  \cite{Beyer_2022_CVPR} introduced the concept of function matching to improve the transfer of knowledge from teacher to student by making both see the consistent input and training the students using a long training schedule. Shen \etal \cite{shen2021fast} used the relabelling technique of Yun \etal \cite{yun2021re} along with data augmentation to boost the performance of the student network in an efficient manner. Li \etal \cite{li2021smile} and Yang \etal \cite{yang2022mixskd} distilled the student using self-knowledge fused with mixup at the feature level, i.e. the feature maps were mixed using the mixup technique. Malinin \etal \cite{malinin2019ensemble} proposed an ensemble approach to improve the calibration of the model by using the prior networks \cite{malinin2018predictive} in the distillation framework.
There has been a lot of research in KD \cite{zhao2022decoupled, Beyer_2022_CVPR, zhang2022confidence} to improve the accuracy of the student in different distillation modes.
\begin{figure}[t]
  \centering
  \begin{subfigure}{0.15\textwidth}
  \includegraphics[scale=0.14]{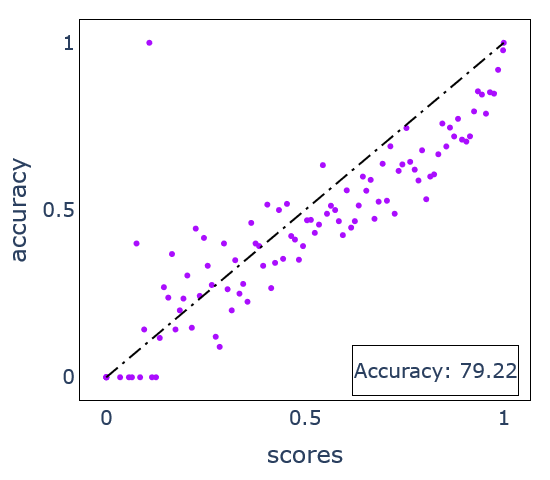}
  \caption{Teacher}
  \label{fig:short-a-teacher}
  \end{subfigure}
  \centering
  \begin{subfigure}{0.15\textwidth}
  \includegraphics[scale=0.14]{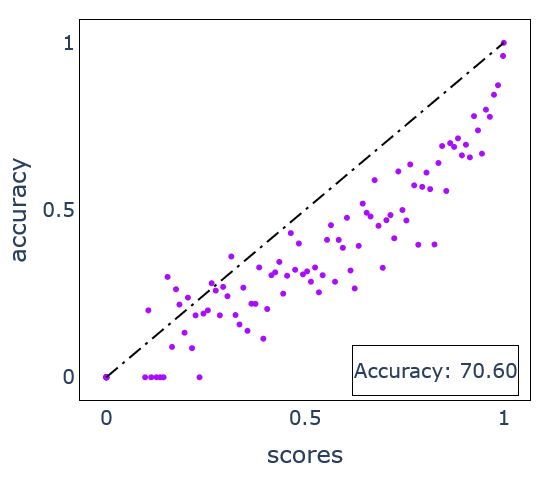}
    \caption{Student}
    \label{fig:short-b-student}
  \end{subfigure}
  \centering
  \begin{subfigure}{0.15\textwidth}
  \includegraphics[scale=0.14]{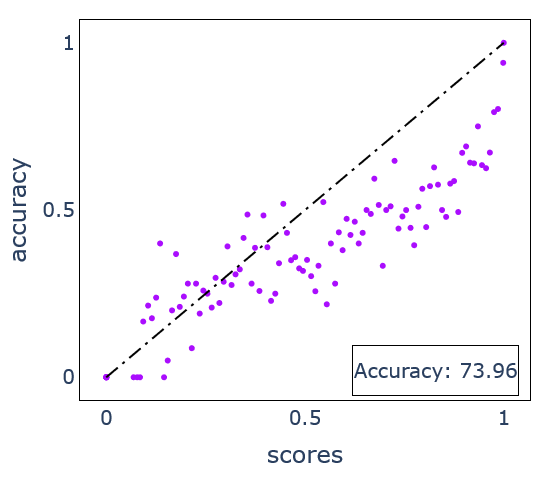}
    \caption{KD}
    \label{fig:short-d-kd_batchmean}
  \end{subfigure}
  
  \caption{(a), (b) and (c) show the scatter plots of accuracy vs confidence for teacher, student and student distilled using conventional KD on CIFAR-100 dataset respectively. A well calibrated model will have most of its density along the $x=y$ line. We observe that (i) the teacher is not calibrated but accurate, (ii) the student model trained from scratch is poorly calibrated and less accurate than the teacher, (iii) by performing distillation, we get a student that is still not well calibrated.}
  \label{fig:intro-cali}
\end{figure}
\subsection{Knowledge distillation with data augmentation}
Wang \etal \cite{wang2022what} demonstrated the role of data augmentation in KD. They passed both the original and the augmented data through the teacher network while training the student network. Zhao \etal \cite{zhao2021similarity} proposed similarity transfer for KD. They use mixup technique to generate virtual samples forming a new dataset, which distills knowledge from the teacher to the student. Zhang \etal \cite{zhang2022cekd} proposed an augmentation-based distillation technique that focuses on improving the accuracy of the student network. They perform cross-distillation using mixup and CutMix augmentation techniques. Xu \etal \cite{xu2020computation} proposed a computationally efficient KD by combining uncertainty and mixup with KD. 
It was discussed in \cite{9711488} that a better teacher does not necessarily make a better student and there may also be a degradation in the accuracy of the distilled student. Muhammad \etal \cite{muhammad2021mixacm} proposed mixup-based robustness transfer via distillation of activated channel maps (MixACM) that makes the student robust by distilling knowledge from the adversarially trained robust teacher using mixup augmentation. 

The advantage of using data augmentation and KD in an integrated framework for calibrating the student network is an open problem. Researchers have mainly focused on improving the accuracy of the student model while overlooking the calibration aspect. Our approach addresses this gap by keeping its prime focus on improving the calibration of the student network without using a robust teacher. To the best of our knowledge, none of the previous works addresses the problem of distilling a calibrated student from an uncalibrated teacher in the image domain. In cases where an improvement in calibration of the student has been reported, the teacher for such approaches was either adversarially trained or a certain technique was used to make the teacher robust. Subsequently, a calibrated teacher resulted in a calibrated student with standard KD. This differs from our approach as we do not make teachers calibrated or robust and use it the way it is. 

\section{Proposed Method}
In this section, we first revisit the concept of distillation followed by data augmentation techniques and then present the generalised framework of our approach.
\textit{Notation:} Let $S$ be the student model and $T$ be the teacher model. This teacher model is pre-trained on the data $\mathcal{D}$. Let $f_t$ and $f_s$ represent the output (logits) of the teacher and student network respectively. Let $X$ represent the original data and $y$ represent its corresponding label.
\begin{figure}[t]
  \centering
  \begin{subfigure}{0.22\textwidth}
  \includegraphics[scale=0.20]{figures/ece_mean_batchmean/kd_batchmean.png}
  \caption{KD}
  \label{fig:batchmean_kd_1}
  \end{subfigure}
  \centering
  \begin{subfigure}{0.22\textwidth}
  \includegraphics[scale=0.20]{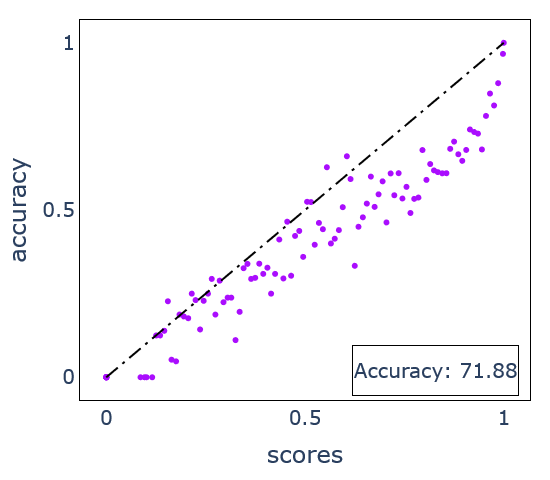}
    \caption{Scaled KD}
    \label{fig:scaled_kd_1}
  \end{subfigure}
  \caption{(a) and (b) show the calibration plots for vanilla KD and scaled KD with $\alpha =0.5$. Scaled KD has a better calibration than vanilla KD but with decreased accuracy. }
  \label{fig:KD-scaled_KD}
\end{figure}
\subsection{Knowledge distillation}
The primary goal of KD is to distill the dark knowledge from the teacher network into the student network by comparing the temperature-scaled softmax logits of the teacher and student. The loss term generally consists of two losses, KL divergence loss ($\mathcal{L}_{div}$) and a task-specific loss ($\mathcal{L}_{task}$).
\begin{equation}\label{kl_basic}
    \mathcal{L}_{div} = \sum_{x_i \in X}\text{KL}[f_s(x_i)||(f_t(x_i)]
\end{equation}
\begin{equation}\label{nll_basic}
    \mathcal{L}_{task} = \sum_{i=1}^{N} -y_i\text{ log } f_s(x_i)
\end{equation}
where $N$ is the total number of samples in the dataset. KL loss is the Kullback-Leibler Divergence score which describes the difference between the probability distribution of soft labels of teacher $T$ and student $S$. In conventional KD \cite{hinton2015distilling}, the KL-divergence loss and the task-specific loss are linearly combined as follows:
\begin{equation}
    \mathcal{L}_{HKD} = \alpha *  \mathcal{L}_{div} + (1- \alpha) * \mathcal{L}_{task}
\end{equation}
where $\alpha$ is a balancing factor. Generally, the value of $\alpha$ is kept high to ensure high knowledge transfer from the teacher to the student, however, doing so makes the student network highly uncalibrated as shown in \cref{fig:intro-cali}. Therefore, we scale the KL-divergence loss by taking mean across the dimension of the logits (i.e. dividing it by the logit size, which is same as number of classes). Using this scaled KL-divergence loss we observe a improvement in the calibration and a drop in accuracy, as shown in \cref{fig:KD-scaled_KD}. We encounter a trade-off between accuracy and calibration error in the framework of KD\footnote{For detailed experiments on KD (without scaling) and without using ground truth, please refer to the supplementary material.}. Our approach leverages the distillation framework to take care of the accuracy while improving the calibration with the help of data augmentation techniques.

\subsection{Distillation with Augmentation: Integrated Framework for Calibration}
In this section, we discuss a generic framework that makes our approach work on any distillation technique with any augmentation technique. Let $A_{KD}$ be a distillation scheme having its loss as $l_{A}$ and $G_{aug}$ be an augmentation scheme that generates $x_G$ samples using the original data samples $x$ and having its own loss as $l_G$. We replace the task-specific loss with the loss corresponding to the augmentation scheme. If the augmentation scheme does not have its own loss, we use the standard cross entropy loss as a task-specific loss. In our framework, we pass the original samples through both the teacher and student networks and obtain the logits/features corresponding to the original data as,
\begin{equation}\label{kd_generic}
    \mathcal{L}_{KD} = l_A(f_t(x),f_s(x))
\end{equation}
We compute the distillation loss $\mathcal{L}_{KD}$ between the outputs of teacher and student using the distillation-specific loss function $l_A$ as shown in \cref{kd_generic}. Our framework is independent of the nature of this loss, i.e., whether it is applied at the feature level or logit level.
Simultaneously, we pass the augmented samples through the student network (only) and apply the augmentation-specific loss.
\begin{equation}
    \mathcal{L}_{Aug} = l_{G}(y,z_s^{aug})
\end{equation}
Here $l_{G}$ is the augmentation-specific loss function, $y$ is the true label corresponding to the augmented sample $x_G$ and $z_s^{aug}=f_s(x_G)$ is the logit vector corresponding to the augmented input $x_G$. As the teacher is uncalibrated, there is no benefit of passing the augmented input through it. The overall loss is represented as follows:
\begin{equation}
    \mathcal{L} = \alpha_A * \mathcal{L}_{KD} + (1-\alpha_A) * \mathcal{L}_{Aug}
\end{equation}
The balancing factor $\alpha_A$ is a hyper-parameter corresponding to the $A_{KD}$ distillation scheme. This hyper-parameter regulates the trade-off between the calibration and accuracy of the student model. 

\subsection{Improving calibration w/o accuracy degradation}
Calibration implies that the model is generalized and does not give overconfident predictions. Various data-augmentation schemes like mixup and CutMix are theoretically proven or empirically shown in the literature to improve the calibration and accuracy of the standalone networks. In our framework, the student learns the knowledge from two sources: distillation and augmentation. Distillation loss targets accuracy, while augmentation-specific loss targets the calibration of the student model. In conventional KD, higher weightage is given to the distillation loss, however, to improve the calibration of the student, the augmentation-specific loss needs more weightage. As the teacher is uncalibrated, improvement in calibration is more likely to come from the augmentation. 
\begin{figure}[t]
    \centering
    \includegraphics[width=\linewidth]{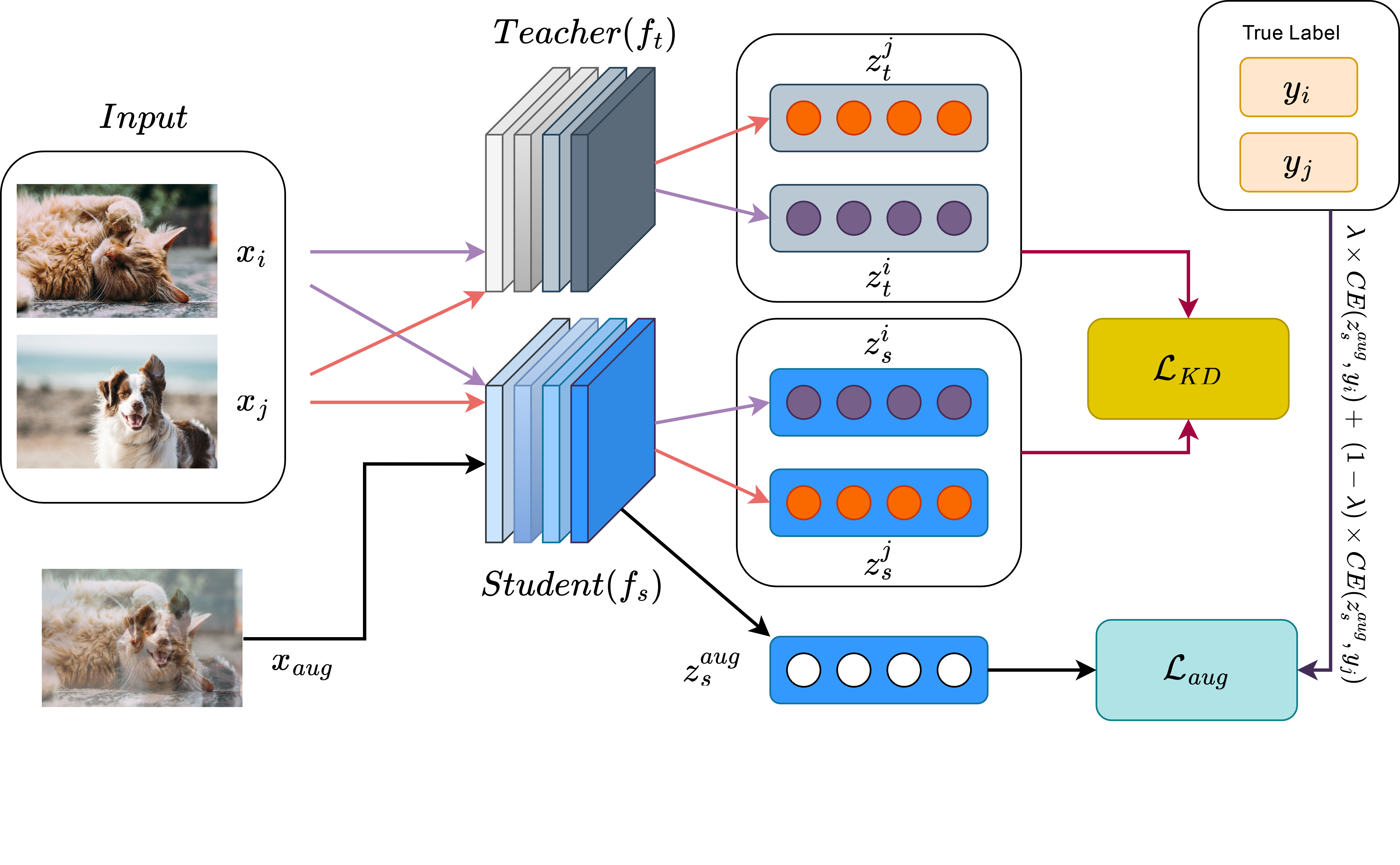}
    \caption{Illustration of our proposed framework. The original image is passed through the teacher $f_t$ and the student $f_s$ network to compute the KL divergence or distillation loss $\mathcal{L}_{KD}$ between the soft logits of teacher and student. Since the teacher is uncalibrated, the augmented image $x_{aug}$ is passed through the student network only. The augmentation-based loss $\mathcal{L}_{aug}$ is then applied to the logits $z^{aug}_s$ and true labels. }
    
    \label{fig:our approach-mixup}
\end{figure}
\subsubsection*{Calibrated distillation using mixup}
Mixup is a data-augmentation technique developed by \cite{zhang2018mixup}. It is a linear interpolation of the images and labels to generate the new augmented data. Thulasidasan \etal\cite{thulasidasan2019mixup} acknowledged the calibration and generalization property of mixup. Zhang \etal \cite{zhang2022and} theoretically proved the improvement in calibration metrics when the model is trained using mixup. In addition, mixup is also shown to reduce the overconfidence error of the model.
Consider two random samples from the dataset, $(x_i,y_i)$ and $(x_j,y_j)$. Mixup generates a new data point in the vicinity of the original data distribution by using these two data points controlled by hyperparameter ``$\lambda$''.  Mathematically, mixup is defined as follows:
\begin{align} 
x_{mix} = \lambda * x_i + (1-\lambda) * x_j \label{mixup_eq1}\\
    y_{mix} = \lambda * y_j + (1-\lambda) * y_j \label{mixup_eq2}
\end{align}
The empirical Dirac delta distribution centered at $(x_i, y_i)$ 
\begin{equation}
    P_\delta(x,y) = \frac{1}{n} \sum_{i}^{n} \delta(x=x_i,y=y_i)
\end{equation}
is replaced with the empirical vicinal distribution
\begin{equation}
    P_v(x_{mix},y_{mix}) = \frac{1}{n} \sum_i^n \nu(x_{mix},y_{mix}| x_i,y_i)
\end{equation}
where $(x_{mix},y_{mix})$ is the mixup generated data point, and $\nu$ is the vicinal distribution that gives the probability of the new sample in the vicinity of the original data distribution. This vicinity distribution gives the probability $P_v$ of finding an augmented sample.
Usually $\lambda$ is sampled from a beta distribution $\beta(a,a)$ where $a \in (0,\infty)$. We will not use soft label $y_{mix}$ in future operations; therefore, we skip this computation, instead we use hard labels.
The mixup loss is defined as follows:
\begin{equation}\label{eq11}
    \mathcal{L}_{CE} = - [\lambda * y_i * log(f_s(x_{mix})) + (1-\lambda)* y_j * log(f_s(x_{mix}))] 
\end{equation}
where $y_i$ and $y_j$ are the true labels corresponding to the image $x_i$ and $x_j$. 
Since mixup uses two images to generate an augmented image, we modify the distillation loss as follows:
\begin{equation}\label{eq12}
    \mathcal{L}_{KD} = \frac{1}{n_c} (KL[f_t(x_i)||f_s(x_i)] + KL[f_t(x_j)||f_s(x_j)])
\end{equation}
where $n_c$ is the number of classes. The distillation objective function takes the following form:
\begin{align}\label{loss_eq_kd}
     \mathcal{L}=\alpha \mathcal{L}_{KD} + (1-\alpha) \mathcal{L}_{CE}
\end{align}
The hyper-parameter $\alpha$ signifies the amount of knowledge to be gained by the student from the teacher and by itself via training using augmented data. Our integrated approach using mixup is shown in \cref{fig:our approach-mixup}. Similar to mixup, we apply our technique to calibrate student models using cutout and CutMix\footnote{For details and implementation please refer to supplementary material}.

\section{Experiments}
We present our experiments on the calibration of the distilled classifiers. We take mean of the KL-divergence loss by dividing it by the size of the logit vector and refer to this scaled version as conventional KD unless stated otherwise. We conduct several experiments to observe the student network calibration using the popular data augmentation techniques such as cutout, mixup, and CutMix. We use datasets such as CIFAR-10/100, CINIC-10, and TinyImageNet for image classification experiments and report the calibration metrics (ECE and OE) corresponding to best validation accuracy. For understanding the robustness of our approach we use CIFAR-100C dataset\footnote{We perform experiments on OOD data also and include our findings in the supplementary material.}.
\subsection{Calibration Metrics}
In the experiments, we use two calibration metrics namely \textbf{ECE} (expected calibration error) and \textbf{OE} (overconfidence error) defined as follows:
\begin{equation}
    \textbf{ECE} = \sum^B_{b=1} \frac{n_b}{N} | acc(b) - conf(b) |
\end{equation}
\begin{equation}
    \textbf{OE} = \sum^B_{b=1} \frac{n_b}{N} [conf(b) \times max(conf(b)-acc(b),0)]
\end{equation}
 where $n_b$ is the number of predictions in bin $b$, $N$ is the number of testing samples, $B$ is the total number of bins, $acc(b)$ and $conf(b)$ are the accuracy and average confidence corresponding to bin $b$.
We follow Thulasidasan \etal \cite{thulasidasan2019mixup} implementation of ECE and OE. 
\begin{figure*}[!t]
  \centering
  \begin{subfigure}{0.157\linewidth}
  \includegraphics[scale=0.15]{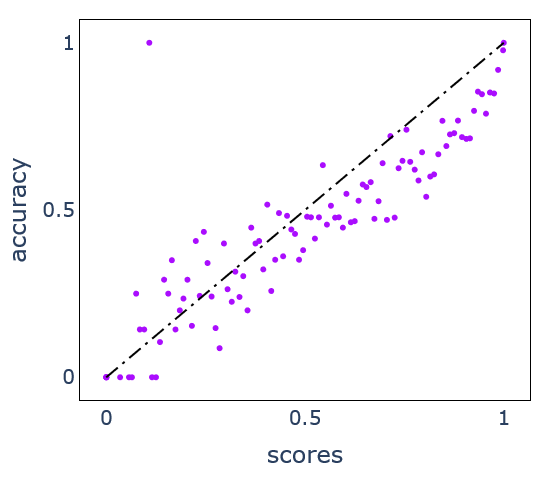}
  \caption{Teacher}
  \label{fig:short-a}
  \end{subfigure}
  \centering
  \begin{subfigure}{0.157\linewidth}
  \includegraphics[scale=0.147]{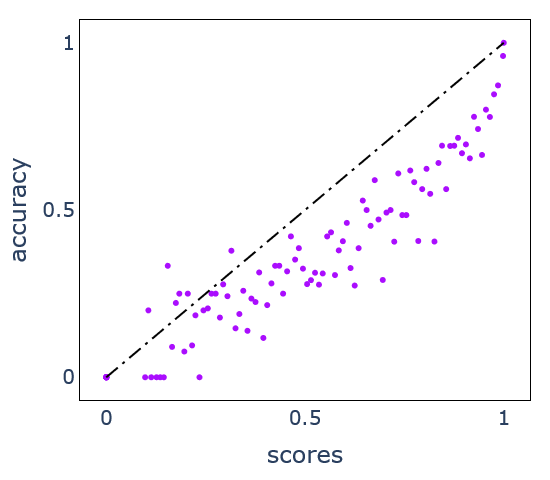}
    \caption{Student}
    \label{fig:short-b}
  \end{subfigure}
  \centering
  \begin{subfigure}{0.157\linewidth}
  \includegraphics[scale=0.147]{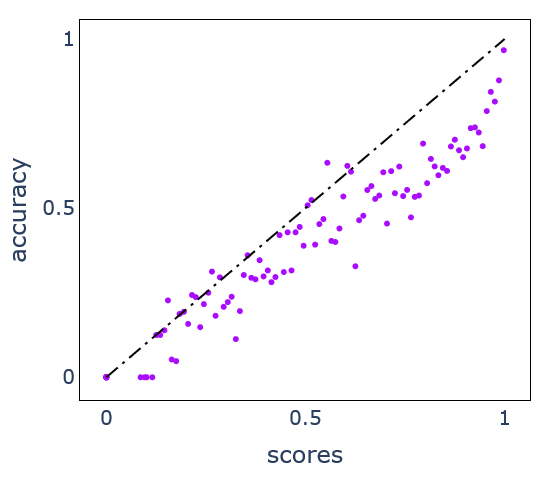}
    \caption{KD}
    \label{fig:short-c}
  \end{subfigure}
  \centering
  \begin{subfigure}{0.157\linewidth}
  \includegraphics[scale=0.147]{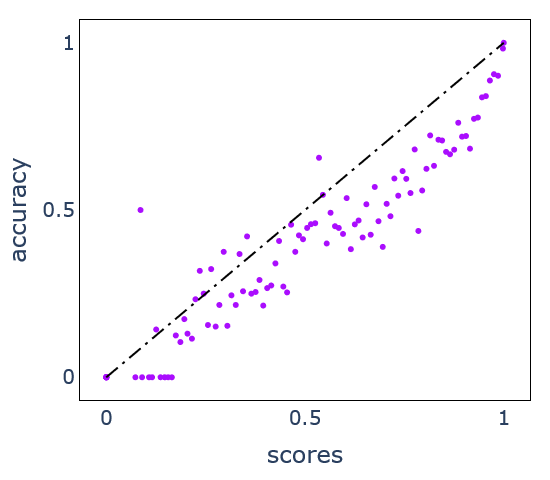}
    \caption{KD~+~cutout}
    \label{fig:short-kd-cutout}
  \end{subfigure}
  \centering
  \begin{subfigure}{0.157\linewidth}
  \includegraphics[scale=0.147]{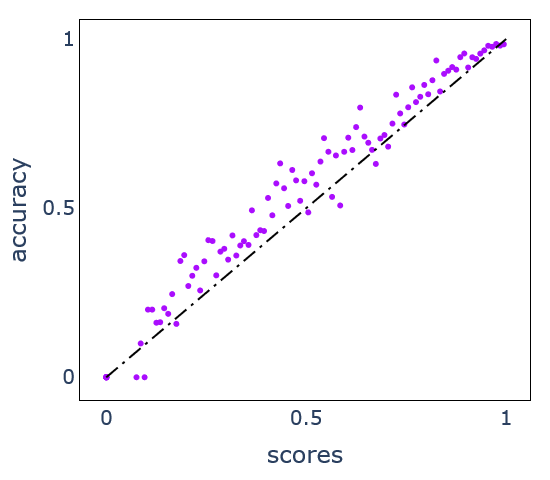}
    \caption{KD~+~mixup}
    \label{fig:short-kd-mixup}
  \end{subfigure}
  \centering
  \begin{subfigure}{0.157\linewidth}
  \includegraphics[scale=0.147]{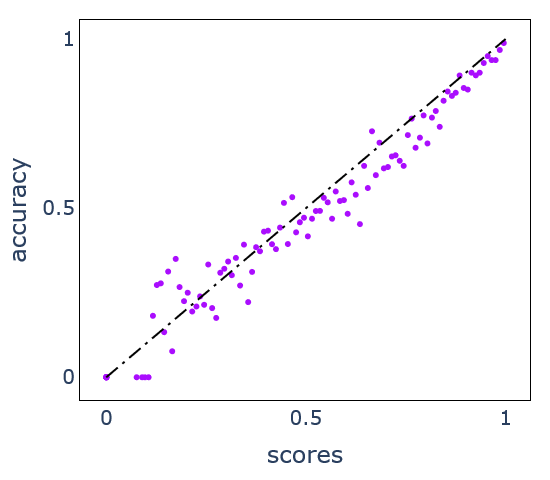}
    \caption{KD~+~CutMix}
    \label{fig:short-kd-cutmix}
  \end{subfigure}
  \caption{Calibration plot for CIFAR-100 dataset. (a), (b) and (c) are identical to Fig. \ref{fig:short-a-teacher}, \ref{fig:short-b-student} and \ref{fig:scaled_kd_1}. They are included here only for completeness. (d), (e) and (f) are calibration plots for KD~+~cutout, KD~+~mixup and KD~+~CutMix respectively. The teacher is ResNet32x4 and the student is ShuffleNetV1. The graph shows a scatterplot of accuracy vs confidence. We observe that KD~+~mixup and KD~+~CutMix shows a considerable improvement in the calibration of the student.}
  \label{fig:calibration-plots}
\end{figure*}
\begin{table*}
\centering
\begin{tabular}{cccccccccc}
\toprule
Teacher/                                                                     & Metrics & Teacher & Student & KD     & RKD    & CRD    & KD +  & KD +         & KD +        \\
Student&&&&&& & cutout & mixup & CutMix\\
\midrule
\multirow{3}{*}{\begin{tabular}[c]{@{}c@{}}WRN-40-2/\\ ShuffleNetV1\end{tabular}}   & Acc     & 77.09   & 70.58   & 71.05  & 73.41  & 75.28  & 72.46     & 74.65           & 74.94           \\
                                                                                    & ECE     & 0.103  & 0.121  & 0.108 & 0.126 & 0.106  & 0.090    & {\ul 0.049}    & \textbf{0.046} \\
                                                                                    & OE      & 0.085  & 0.093  & 0.084 & 0.100 & 0.087 & 0.067    & \textbf{0.002} & {\ul 0.031}    \\
\midrule
\multirow{3}{*}{\begin{tabular}[c]{@{}c@{}}ResNet32x4/\\ ShuffleNetV2\end{tabular}} & Acc     & 79.21   & 73.19   & 72.58  & 73.68  & 75.14  & 74.05     & 75.25           & 76.64           \\
                                                                                    & ECE     & 0.068  & 0.104  & 0.104 & 0.122  & 0.105 & 0.087    & \textbf{0.042} & {\ul 0.061}    \\
                                                                                    & OE      & 0.052  & 0.083  & 0.082 & 0.098 & 0.087 & 0.066    & \textbf{0.003} & {\ul 0.044}    \\
\midrule
\multirow{3}{*}{\begin{tabular}[c]{@{}c@{}}ResNet50/\\ MobileNetV2\end{tabular}}    & Acc     & 78.98   & 63.2    & 63.92  & 63.81  & 67.53  & 64.03     & 65.94           & 67.32           \\
                                                                                    & ECE     & 0.104  & 0.169  & 0.135 & 0.167 & 0.134 & 0.097    & {\ul 0.051}    & \textbf{0.037} \\
                                                                                    & OE      & 0.086   & 0.129  & 0.101  & 0.128 & 0.103 & 0.068    & \textbf{0.001} & {\ul 0.019}    \\
\midrule
\multirow{3}{*}{\begin{tabular}[c]{@{}c@{}}ResNet50/\\ VGG8\end{tabular}}           & Acc     & 78.98   & 70.3    & 70.79  & 71.4   & 73.27  & 71.47     & 71.94           & 72.3            \\
                                                                                    & ECE     & 0.104  & 0.122  & 0.091 & 0.103 & 0.101  & 0.087    & {\ul 0.050}    & \textbf{0.031} \\
                                                                                    & OE      & 0.086   & 0.096  & 0.070 & 0.077 & 0.080 & 0.063    & \textbf{0.002} & {\ul 0.011}    \\
\midrule
\multirow{3}{*}{\begin{tabular}[c]{@{}c@{}}ResNet32x4/\\ ShuffleNetV1\end{tabular}} & Acc     & 79.21   & 70.58   & 71.86  & 71.96  & 73.98  & 73.23     & 74.68           & 74.63           \\
                                                                                    & ECE     & 0.068  & 0.121  & 0.096 & 0.130 & 0.110 & 0.087    & {\ul 0.051}    & \textbf{0.036} \\
                                                                                    & OE      & 0.052   & 0.093  & 0.077 & 0.103  & 0.089  & 0.065     & \textbf{0.001}  & {\ul 0.023}    \\
\midrule
\multirow{3}{*}{\begin{tabular}[c]{@{}c@{}}VGG13/\\ MobileNetV2\end{tabular}}       & Acc     & 76.36   & 63.2    & 64.13  & 63.44  & 67.35  & 64.03     & 66.48           & 66.44           \\
                                                                                    & ECE     & 0.073  & 0.169  & 0.130 & 0.167 & 0.114 & 0.103    & {\ul 0.047}    & \textbf{0.037} \\
                                                                                    & OE      & 0.057  & 0.129  & 0.097 & 0.127 & 0.086 & 0.071    & \textbf{0.002}  & {\ul 0.022}   \\
\bottomrule
\end{tabular}
\caption{Experiment result on CIFAR-100 dataset. The table shows the result of our framework on six teacher-student pairs. We observe a significant improvement in the calibration of the student for all the teacher-student pairs considered for CIFAR-100 dataset. The best scores are highlighted in bold and second best are underlined.}
\label{cifar100-table1}
\end{table*}
\begin{table*}[t]
\centering

\begin{tabular}{cccccccccc}
\toprule
Teacher   & \multicolumn{3}{c}{VGG13}                   & \multicolumn{3}{c}{ResNet50}                 & \multicolumn{3}{c}{VGG13}                    \\
Student   & \multicolumn{3}{c}{MobileNetV2}             & \multicolumn{3}{c}{MobileNetV2}              & \multicolumn{3}{c}{VGG11}                    \\
\midrule
Metrics   & Accuracy & ECE             & OE             & Accuracy & ECE             & OE              & Accuracy & ECE             & OE              \\
\midrule
Teacher   & 93.92    & 0.039          & 0.034          & 95.51    & 0.027          & 0.024          & 93.92    & 0.039          & 0.034           \\
Student   & 95.01    & 0.031          & 0.027         & 95.01    & 0.031          & 0.027          & 92.4     & 0.047          & 0.041          \\
KD        & 95.19    & 0.034          & 0.031         & 95.26    & 0.035          & 0.031          & 92.84    & 0.050          & 0.044          \\
\midrule
KD~+~cutout & 96.25    & 0.025           & 0.022         & 96       & \textbf{0.021} & {\ul 0.018}    & 93.38    & 0.040          & 0.035          \\
KD~+~mixup  & 95.83    & {\ul 0.021}    & \textbf{0.013} & 95.75    & 0.023          & \textbf{0.012} & 93.35    & {\ul 0.028}    & \textbf{0.015} \\
KD~+~CutMix & 96.3     & \textbf{0.019} & {\ul 0.014}   & 96.22    & {\ul 0.021}    & 0.018          & 93.64    & \textbf{0.022} & {\ul 0.016}   \\
\bottomrule
\end{tabular}
\caption{Experiments on CIFAR-10 dataset: Our approach improves the calibration error of the student network as compared to traditional KD. Among various augmentation techniques, CutMix augmentation integrated using our approach has distilled a calibrated as well as accurate student.}
\label{Cifar10-table}
\end{table*}
\subsection{Setup}
For CIFAR-10/100 and CINIC-10 dataset experiments, we use SGD optimizer having momentum of 0.9, weight decay of $5 \times 10^{-4}$ and initial learning rate of 0.1 (0.01 for MobileNet, ShuffleNet architectures). Similar to \cite{tian2019crd}, we train networks for 240 epochs with batch size of 64 for CIFAR10/100 and 256 for CINIC-10 datasets. We use CosineAnnealingLR as a learning rate scheduler with $T_{max}=$240.
For TinyImageNet dataset experiments, we use the same optimizer except a weight decay of $10^{-4}$. We train the networks for 200 epochs with batch size of 128 and scheduled the learning rate using MutliStepLR having milestones at 150 and 180 epochs and decay factor of 0.1. For RKD and CRD experiments, we use author provided hyper-parameters for their respective loss functions. For all the experiments, conventional KD’s Temperature, mixup $\beta$ distribution parameter ($a$), and loss balancing factor ($\alpha$)  are fixed as 50, 0.4, and 0.5 respectively unless otherwise stated. For the experiments involving cutout augmentation technique, the number of holes is 1 and size of masked region is 16 $\times$ 16. We use the standard hyper-parameters for CutMix augmentation technique (number of holes =1, CutMix probability=0.5). In all the experiments, standard augmentations like random crop, horizontal flip, and normalization are applied to the images. 
\subsection{Image Classification}
\subsubsection*{CIFAR-100}
The CIFAR-100 dataset consists of 60,000 RGB images of size 32 $\times$ 32. The dataset has 100 classes where each class consists of 500 train images and 100 test images. 
In our experiments , we use six (teacher, student) pairs as shown in \cref{cifar100-table1}.
For CIFAR-100 experiments, the teacher and student networks have different architectures. We refer scaled KD as KD in all our experiments as scaling helps in calibration (refer \cref{fig:KD-scaled_KD}). We use KD (scaled), RKD \cite{park2019relational} and CRD \cite{tian2019crd} as baselines for comparison. \cref{cifar100-table1} shows that the students distilled through our integrated framework are better calibrated with comparable accuracy and reduced overconfidence. Note that unlike the well-engineered and advance distillation approaches, RKD and CRD, we use conventional KD while distilling a calibrated student through our approach. Later in \cref{Extension to other KD techniques}, we show that RKD and CRD can also be improved through proposed framework.

Out of the 3 augmentation techniques considered in our experiments, CutMix and mixup calibrates the student model better than cutout. CutMix, which takes the advantage of both cutout and mixup does better than mixup in most cases. Our approach has reduced the ECE of the student model up to 5 times when compared with that of the student trained from scratch. In all the teacher-student configurations, it is observed that the student model has an improved calibration (ECE) than the corresponding teacher network. Mixup \cite{thulasidasan2019mixup} targets the overconfidence of the model, therefore KD~+~mixup drastically reduces the OE of the model which makes it suitable for high risk domains. 
The calibration plots for teacher-student pair: ResNet32x4 and ShuffleNetV1 are shown in \cref{fig:calibration-plots}. We observe that KD and KD~+~cutout slightly improves the calibration of the student network while KD~+~mixup and KD~+~CutMix shows significant improvement in the calibration by aligning the confidence of the model with the accuracy. 
\begin{table*}[!t]
\centering
\begin{tabular}{cccccccccc}
\toprule
Teacher   & \multicolumn{3}{c}{VGG13}                    & \multicolumn{3}{c}{ResNet50}                 & \multicolumn{3}{c}{VGG13}                    \\
Student   & \multicolumn{3}{c}{MobileNetV2}              & \multicolumn{3}{c}{MobileNetV2}              & \multicolumn{3}{c}{VGG11}                    \\
\midrule
Metrics   & Accuracy & ECE             & OE              & Accuracy & ECE             & OE              & Accuracy & ECE             & OE              \\
\midrule
Teacher   & 84.68    & 0.105           & 0.095          & 88.2     & 0.085          & 0.078          & 84.68    & 0.105           & 0.095          \\
Student   & 86.43    & 0.092          & 0.083          & 86.43    & 0.092          & 0.083          & 81.62    & 0.125          & 0.112          \\
KD        & 87.05    & 0.094          & 0.087          & 86.95    & 0.101           & 0.094          & 82.54    & 0.126          & 0.115          \\
\midrule
KD~+~cutout & 88.1     & 0.075          & 0.067          & 88.04    & 0.075          & 0.067          & 83.22    & 0.106          & 0.093          \\
KD~+~mixup  & 87.24    & \textbf{0.044} & {\ul 0.035}    & 87.64    & {\ul 0.048}    & {\ul 0.041}    & 82.31    & {\ul 0.093}    & {\ul 0.080}    \\
KD~+~CutMix & 87.75    & {\ul 0.045}    & \textbf{0.003} & 87.69    & \textbf{0.030} & \textbf{0.018} & 83.12    & \textbf{0.025} & \textbf{0.011} \\
\bottomrule
\end{tabular}
\caption{Experiments on CINIC-10 dataset. Our approach outperforms the conventional KD in the calibration metrics by reducing the ECE to approximately half when compared with the student trained from scratch.}
\label{cinic-10-results}
\end{table*}
\subsubsection*{CIFAR-10 and CINIC-10}
The CIFAR-10 consists of 60,000 RGB images of size 32 $\times$ 32. The dataset has 10 classes where each class consists of 5000 train images and 1000 test images.   We perform experiments on various architectures to show the effectiveness of our approach in calibrating the student network. The results of our experiments are shown in Table \ref{Cifar10-table}. Since the dataset is small, we observe a good ECE for teachers. In the (VGG13, MobileNetV2) combination, we observe improvement in calibration when the teacher has higher ECE than student. In other combinations, the teacher network has lower ECE compared to the student. Our approach outperforms the conventional KD in terms of ECE and OE. The student network distilled using our approach is well-calibrated and even more accurate. Among the augmentations considered, KD~+~CutMix approach results in comparatively better calibration similar to the previous experiments. We observe that the KD~+~mixup, similar to previous experiments, decreases the overconfidence by a large factor, and KD~+~cutout approach also gives a better-calibrated student when compared with the student (trained from scratch) and KD.
\paragraph{}We extend our observation to CINIC-10 dataset which is downsampled from both CIFAR-10 and ImageNet. It consists of 270,000 colored images of size 32x32. The dataset contains training, validation and test set with each set having 90,000 images. The results for CINIC-10 are shown in \cref{cinic-10-results}. We observe a similar trend as in CIFAR-10 with improvement in the calibration of the student model trained using our framework. For CINIC-10, CutMix approach has performed comparatively better in terms of calibration.
\begin{table}[!t]
\centering

\begin{tabular}{lccc}
\toprule
Metrics & \multicolumn{1}{c}{Accuracy} & \multicolumn{1}{c}{ECE} & OE \\ \midrule
Teacher & \multicolumn{1}{c}{57.28} & \multicolumn{1}{c}{0.140} & 0.100 \\
Student & \multicolumn{1}{c}{55.76} & \multicolumn{1}{c}{0.146} & 0.103 \\
KD & \multicolumn{1}{c}{56.22} & \multicolumn{1}{c}{0.103} & 0.072 \\ \midrule
KD~+~cutout & \multicolumn{1}{c}{58.6} & \multicolumn{1}{c}{0.102} & 0.070\\
KD~+~mixup & \multicolumn{1}{c}{58.11} & \multicolumn{1}{c}{{\ul 0.074}} & \textbf{0.004} \\

KD~+~CutMix & \multicolumn{1}{c}{60.99} & \multicolumn{1}{c}{\textbf{0.049}} & {\ul 0.026} \\
\bottomrule
\end{tabular}
\caption{Experiments on TinyImageNet Dataset. We report the calibration metrics of the best validation accuracy of the model obtained through training.}
\label{TinymageNet}\end{table}
\begin{table}[!t]
\centering
\begin{tabular}{lccc} 
\\ \toprule
Metrics               & Accuracy   & ECE    & OE \\
\midrule
RKD-DA                & 72.81 & 0.113 &  0.089  \\
RKD-DA + cutout & 73.7 & 0.102 & 0.080 \\ 
RKD-DA + mixup  & 75.87  & \textbf{0.041} &  {\ul 0.002}  \\
RKD-DA + CutMix & 74.81 & {\ul 0.046} &  \textbf{0.001}  \\
\midrule
CRD            & 75.03 & 0.083 &  0.064  \\
CRD + cutout    &    75.84   &   0.080     &  0.063  \\
CRD + mixup     & 76.56 & {\ul 0.065} &  \textbf{0.002}  \\
CRD + CutMix    &    76.41   &   \textbf{0.046}     &  {\ul 0.015}  \\

\bottomrule
\end{tabular}
\caption{Experiment results of RKD and CRD distillation techniques on CIFAR-100.}
\label{extension}
\end{table}
\begin{table*}[!t]
\centering
\setlength{\tabcolsep}{4pt}
\begin{tabular}{ccccccccccccc}
\toprule
\multicolumn{1}{l}{\textbf{}} & \multicolumn{3}{c}{Frost}                    & \multicolumn{3}{c}{Snow}                     & \multicolumn{3}{c}{Gaussian Blur}           & \multicolumn{3}{c}{Motion Blur}              \\
\midrule
Metrics                       & Acc & ECE             & OE              & Acc & ECE             & OE              & Acc & ECE            & OE              & Acc & ECE             & OE              \\
\midrule
Teacher                       & 51.77    & 0.177          & 0.132          & 57.33    & 0.128          & 0.095          & 54.51    & 0.176          & 0.132          & 57.9     & 0.140          & 0.105          \\
Student                       & 51.36    & 0.200          & 0.146          & 53.53    & 0.185          & 0.134          & 50.09    & 0.224          & 0.163          & 51.17    & 0.212          & 0.155          \\
KD                            & 51.03    & 0.188          & 0.139          & 53.56    & 0.166          & 0.121           & 51.33    & 0.183         & 0.132          & 52.99    & 0.175           & 0.128          \\
\midrule
KD~+~cutout                     & 44.91    & 0.204          & 0.141          & 51.33    & 0.166          & 0.115          & 47.88    & 0.198         & 0.135          & 51.17    & 0.181          & 0.128          \\
KD~+~mixup                      & 56.44    & \textbf{0.049} & \textbf{0.018} & 60.24    & \textbf{0.037} & \textbf{0.007} & 51.38    & \textbf{0.076} & \textbf{0.036} & 53.64    & \textbf{0.049} & \textbf{0.022} \\
KD~+~CutMix                     & 49.97    & {\ul 0.132}    & {\ul 0.090}    & 55.82    & {\ul 0.091}    & {\ul 0.061}    & 51.41    & {\ul 0.132}   & {\ul 0.086}    & 52.26    & {\ul 0.127}     & {\ul 0.086}   \\
\bottomrule
\end{tabular}

\caption{Experiment results on CIFAR-100C. We evaluated our models trained on CIFAR-100 dataset on the corrupted data. We observe a better calibration for the corrupted dataset.}
\label{Cifar100c}
\end{table*}
\subsubsection*{TinyImageNet}
TinyImageNet is a subset of the ImageNet dataset. It consists 200 class images of resolution 64x64x3 with a training set consisting of 100,000 images and a validation  set consisting of 10,000 images. In this experiment, we consider ResNet-34 and ResNet18 as teacher and student respectively. The results on the validation set are reported in \cref{TinymageNet}. The teacher and student have high ECE and OE, therefore are poorly calibrated. We observe an considerable improvement in the calibration of the student network for all augmentations considered. KD~+~mixup approach has reduced the overconfidence error by a factor of 50 and KD~+~CutMix reduced the ECE by nearly one-third when compared with the student trained from scratch. The accuracy of models distilled using our mixup and cutmix approach has surpassed the accuracy of the teacher which is an additional benefit. This shows the effectiveness of our framework even on a TinyImageNet which is considerably a large and complex dataset.

\subsection{Testing Robustness}
\subsubsection*{CIFAR-100C}
CIFAR-100C is a corrupted dataset of CIFAR-100 with several noises. It consists of 15 different corruption schemes, with each corruption having 5 severity levels. Each corruption set consists of 50,000 test images. In this experiment we consider four corruption schemes, namely frost, snow, gaussian blur and motion blur.
We use ResNet32x4 as teacher and ShuffleNetV1 as student which are trained on original CIFAR-100 to evaluate the robustness of the proposed method. The results are averaged over all severity levels and reported in \cref{Cifar100c}. In this experiment, we measure the accuracy and calibration of the model when it is exposed to noisy inputs. We observe that in case of snow and frost noise, KD~+~mixup outperforms the performance of teacher, student and conventional distillation both in terms of accuracy and calibration. However, the cutout integration does not perform well as the important regions are blacked out i.e. information is lost when we apply cutout augmentation technique and the model finds it difficult for high severity levels thereby decreasing the overall accuracy. KD~+~cutmix also shows am improvement in accuracy, ECE and OE when compared with conventional distillation. This shows that our framework helps in the making the student robust for corrupted images. The choice of data-augmentation and distillation technique has an effect over the robustness of distilled student. In our experiments, we observe that KD~+~mixup surpasses the accuracy and ECE of conventional KD.

\subsection{Extension to other KD techniques}\label{Extension to other KD techniques}
In this section, we explore the integration of other distillation techniques within our framework on CIFAR-100 dataset.
We extend our approach to RKD\cite{park2019relational} and CRD\cite{tian2019crd}. The teacher-student pair for this framework is (WideResNet-40-2, ShuffleNetV1). 
The experiment results are reported in \cref{extension}. We observe an improvement in the calibration metrics of these approaches along with better accuracy. For RKD and CRD, we replace the distillation loss $\mathcal{L}_{KD}$ in equation by DA (distance-angle) and CRD loss respectively. We have not used any KL-divergence loss in this experiment. The experiments are performed on the hyperparameters suggested by the respective authors except balancing factor which is 0.5 in this experiment. We observe that mixup, CutMix and cutout integrated distillation shows a better calibration and accuracy. 
\section{Conclusions}
In this work, we successfully distilled a well-calibrated student from an uncalibrated teacher. Although augmentation techniques have been utilized in past for accuracy improvement while distillation but their benefit on calibration has remained unexplored. We bring forward this advantage and show it with the help of different distillation and different augmentation techniques. Our approach can be easily integrated to any existing or subsequently developed distillation techniques. We used different teacher-student combinations for different experiments to show the variability and generalizability of our proposed method. In future, we can explore a similar approach for data-free knowledge distillation.


\begin{thebibliography}{10}\itemsep=-1pt

\bibitem{ba2014deep}
Jimmy Ba and Rich Caruana.
\newblock Do deep nets really need to be deep?
\newblock {\em Advances in neural information processing systems}, 27, 2014.

\bibitem{Beyer_2022_CVPR}
Lucas Beyer, Xiaohua Zhai, Am\'elie Royer, Larisa Markeeva, Rohan Anil, and
  Alexander Kolesnikov.
\newblock Knowledge distillation: A good teacher is patient and consistent.
\newblock In {\em Proceedings of the IEEE/CVF Conference on Computer Vision and
  Pattern Recognition (CVPR)}, pages 10925--10934, June 2022.

\bibitem{bucilua2006model}
Cristian Buciluǎ, Rich Caruana, and Alexandru Niculescu-Mizil.
\newblock Model compression.
\newblock In {\em Proceedings of the 12th ACM SIGKDD international conference
  on Knowledge discovery and data mining}, pages 535--541, 2006.

\bibitem{chen2020investigation}
Blair Chen, Liu Ziyin, Zihao Wang, and Paul~Pu Liang.
\newblock An investigation of how label smoothing affects generalization.
\newblock {\em arXiv preprint arXiv:2010.12648}, 2020.

\bibitem{das2020empirical}
Deepan Das, Haley Massa, Abhimanyu Kulkarni, and Theodoros Rekatsinas.
\newblock An empirical analysis of the impact of data augmentation on knowledge
  distillation.
\newblock {\em arXiv preprint arXiv:2006.03810}, 2020.

\bibitem{ding2021local}
Zhipeng Ding, Xu Han, Peirong Liu, and Marc Niethammer.
\newblock Local temperature scaling for probability calibration.
\newblock In {\em Proceedings of the IEEE/CVF International Conference on
  Computer Vision}, pages 6889--6899, 2021.

\bibitem{guo2017calibration}
Chuan Guo, Geoff Pleiss, Yu Sun, and Kilian~Q Weinberger.
\newblock On calibration of modern neural networks.
\newblock In {\em International conference on machine learning}, pages
  1321--1330. PMLR, 2017.

\bibitem{guo2020online}
Qiushan Guo, Xinjiang Wang, Yichao Wu, Zhipeng Yu, Ding Liang, Xiaolin Hu, and
  Ping Luo.
\newblock Online knowledge distillation via collaborative learning.
\newblock In {\em Proceedings of the IEEE/CVF Conference on Computer Vision and
  Pattern Recognition}, pages 11020--11029, 2020.

\bibitem{he2016deep}
Kaiming He, Xiangyu Zhang, Shaoqing Ren, and Jian Sun.
\newblock Deep residual learning for image recognition.
\newblock In {\em Proceedings of the IEEE conference on computer vision and
  pattern recognition}, pages 770--778, 2016.

\bibitem{hinton2015distilling}
Geoffrey Hinton, Oriol Vinyals, Jeff Dean, et~al.
\newblock Distilling the knowledge in a neural network.
\newblock {\em arXiv preprint arXiv:1503.02531}, 2(7), 2015.

\bibitem{huang2017densely}
Gao Huang, Zhuang Liu, Laurens Van Der~Maaten, and Kilian~Q Weinberger.
\newblock Densely connected convolutional networks.
\newblock In {\em Proceedings of the IEEE conference on computer vision and
  pattern recognition}, pages 4700--4708, 2017.

\bibitem{kumar2019verified}
Ananya Kumar, Percy~S Liang, and Tengyu Ma.
\newblock Verified uncertainty calibration.
\newblock {\em Advances in Neural Information Processing Systems}, 32, 2019.

\bibitem{li2021smile}
Xingjian Li, Haoyi Xiong, Chengzhong Xu, and Dejing Dou.
\newblock Smile: Self-distilled mixup for efficient transfer learning.
\newblock {\em arXiv preprint arXiv:2103.13941}, 2021.

\bibitem{malinin2018predictive}
Andrey Malinin and Mark Gales.
\newblock Predictive uncertainty estimation via prior networks.
\newblock {\em Advances in neural information processing systems}, 31, 2018.

\bibitem{malinin2019ensemble}
Andrey Malinin, Bruno Mlodozeniec, and Mark Gales.
\newblock Ensemble distribution distillation.
\newblock {\em arXiv preprint arXiv:1905.00076}, 2019.

\bibitem{minderer2021revisiting}
Matthias Minderer, Josip Djolonga, Rob Romijnders, Frances Hubis, Xiaohua Zhai,
  Neil Houlsby, Dustin Tran, and Mario Lucic.
\newblock Revisiting the calibration of modern neural networks.
\newblock {\em Advances in Neural Information Processing Systems},
  34:15682--15694, 2021.

\bibitem{mobahi2020self}
Hossein Mobahi, Mehrdad Farajtabar, and Peter Bartlett.
\newblock Self-distillation amplifies regularization in hilbert space.
\newblock {\em Advances in Neural Information Processing Systems},
  33:3351--3361, 2020.

\bibitem{muhammad2021mixacm}
Awais Muhammad, Fengwei Zhou, Chuanlong Xie, Jiawei Li, Sung-Ho Bae, and
  Zhenguo Li.
\newblock Mixacm: Mixup-based robustness transfer via distillation of activated
  channel maps.
\newblock {\em Advances in Neural Information Processing Systems}, 34, 2021.

\bibitem{muller2019does}
Rafael M{\"u}ller, Simon Kornblith, and Geoffrey~E Hinton.
\newblock When does label smoothing help?
\newblock {\em Advances in neural information processing systems}, 32, 2019.

\bibitem{nixon2019measuring}
Jeremy Nixon, Michael~W Dusenberry, Linchuan Zhang, Ghassen Jerfel, and Dustin
  Tran.
\newblock Measuring calibration in deep learning.
\newblock In {\em CVPR Workshops}, volume~2, 2019.

\bibitem{park2019relational}
Wonpyo Park, Dongju Kim, Yan Lu, and Minsu Cho.
\newblock Relational knowledge distillation.
\newblock In {\em Proceedings of the IEEE/CVF Conference on Computer Vision and
  Pattern Recognition}, pages 3967--3976, 2019.

\bibitem{romero2014fitnets}
Adriana Romero, Nicolas Ballas, Samira~Ebrahimi Kahou, Antoine Chassang, Carlo
  Gatta, and Yoshua Bengio.
\newblock Fitnets: Hints for thin deep nets.
\newblock {\em arXiv preprint arXiv:1412.6550}, 2014.

\bibitem{shen2021fast}
Zhiqiang Shen and Eric Xing.
\newblock A fast knowledge distillation framework for visual recognition.
\newblock {\em arXiv preprint arXiv:2112.01528}, 2021.

\bibitem{stanton2021does}
Samuel Stanton, Pavel Izmailov, Polina Kirichenko, Alexander~A Alemi, and
  Andrew~G Wilson.
\newblock Does knowledge distillation really work?
\newblock {\em Advances in Neural Information Processing Systems},
  34:6906--6919, 2021.

\bibitem{tan2019efficientnet}
Mingxing Tan and Quoc Le.
\newblock Efficientnet: Rethinking model scaling for convolutional neural
  networks.
\newblock In {\em International conference on machine learning}, pages
  6105--6114. PMLR, 2019.

\bibitem{thulasidasan2019mixup}
Sunil Thulasidasan, Gopinath Chennupati, Jeff~A Bilmes, Tanmoy Bhattacharya,
  and Sarah Michalak.
\newblock On mixup training: Improved calibration and predictive uncertainty
  for deep neural networks.
\newblock {\em Advances in Neural Information Processing Systems}, 32, 2019.

\bibitem{tian2019crd}
Yonglong Tian, Dilip Krishnan, and Phillip Isola.
\newblock Contrastive representation distillation.
\newblock In {\em International Conference on Learning Representations}, 2020.

\bibitem{wang2022what}
Huan Wang, Suhas Lohit, Michael~Jeffrey Jones, and Yun Fu.
\newblock What makes a ''good'' data augmentation in knowledge distillation - a
  statistical perspective.
\newblock In Alice~H. Oh, Alekh Agarwal, Danielle Belgrave, and Kyunghyun Cho,
  editors, {\em Advances in Neural Information Processing Systems}, 2022.

\bibitem{wu2021peer}
Guile Wu and Shaogang Gong.
\newblock Peer collaborative learning for online knowledge distillation.
\newblock In {\em Proceedings of the AAAI Conference on Artificial
  Intelligence}, volume~35, pages 10302--10310, 2021.

\bibitem{xu2020knowledge}
Guodong Xu, Ziwei Liu, Xiaoxiao Li, and Chen~Change Loy.
\newblock Knowledge distillation meets self-supervision.
\newblock In {\em European Conference on Computer Vision (ECCV)}, 2020.

\bibitem{xu2020computation}
Guodong Xu, Ziwei Liu, and Chen~Change Loy.
\newblock Computation-efficient knowledge distillation via uncertainty-aware
  mixup.
\newblock {\em arXiv preprint arXiv:2012.09413}, 2020.

\bibitem{yang2022mixskd}
Chuanguang Yang, Zhulin An, Helong Zhou, Linhang Cai, Xiang Zhi, Jiwen Wu,
  Yongjun Xu, and Qian Zhang.
\newblock Mixskd: Self-knowledge distillation from mixup for image recognition.
\newblock {\em arXiv preprint arXiv:2208.05768}, 2022.

\bibitem{yim2017gift}
Junho Yim, Donggyu Joo, Jihoon Bae, and Junmo Kim.
\newblock A gift from knowledge distillation: Fast optimization, network
  minimization and transfer learning.
\newblock In {\em Proceedings of the IEEE conference on computer vision and
  pattern recognition}, pages 4133--4141, 2017.

\bibitem{yun2019cutmix}
Sangdoo Yun, Dongyoon Han, Seong~Joon Oh, Sanghyuk Chun, Junsuk Choe, and
  Youngjoon Yoo.
\newblock Cutmix: Regularization strategy to train strong classifiers with
  localizable features.
\newblock In {\em Proceedings of the IEEE/CVF international conference on
  computer vision}, pages 6023--6032, 2019.

\bibitem{yun2021re}
Sangdoo Yun, Seong~Joon Oh, Byeongho Heo, Dongyoon Han, Junsuk Choe, and
  Sanghyuk Chun.
\newblock Re-labeling imagenet: from single to multi-labels, from global to
  localized labels.
\newblock In {\em Proceedings of the IEEE/CVF Conference on Computer Vision and
  Pattern Recognition}, pages 2340--2350, 2021.

\bibitem{zagoruyko2016paying}
Sergey Zagoruyko and Nikos Komodakis.
\newblock Paying more attention to attention: Improving the performance of
  convolutional neural networks via attention transfer.
\newblock {\em arXiv preprint arXiv:1612.03928}, 2016.

\bibitem{zhang2022confidence}
Hailin Zhang, Defang Chen, and Can Wang.
\newblock Confidence-aware multi-teacher knowledge distillation.
\newblock In {\em ICASSP 2022-2022 IEEE International Conference on Acoustics,
  Speech and Signal Processing (ICASSP)}, pages 4498--4502. IEEE, 2022.

\bibitem{zhang2018mixup}
Hongyi Zhang, Moustapha Cisse, Yann~N. Dauphin, and David Lopez-Paz.
\newblock mixup: Beyond empirical risk minimization.
\newblock In {\em International Conference on Learning Representations}, 2018.

\bibitem{zhang2022cekd}
Ke Zhang, Jin Fan, Shaoli Huang, Yongliang Qiao, Xiaofeng Yu, and Feiwei Qin.
\newblock Cekd: Cross ensemble knowledge distillation for augmented
  fine-grained data.
\newblock {\em Applied Intelligence}, pages 1--11, 2022.

\bibitem{zhang2022and}
Linjun Zhang, Zhun Deng, Kenji Kawaguchi, and James Zou.
\newblock When and how mixup improves calibration.
\newblock In {\em International Conference on Machine Learning}, pages
  26135--26160. PMLR, 2022.

\bibitem{zhang2019your}
Linfeng Zhang, Jiebo Song, Anni Gao, Jingwei Chen, Chenglong Bao, and Kaisheng
  Ma.
\newblock Be your own teacher: Improve the performance of convolutional neural
  networks via self distillation.
\newblock In {\em Proceedings of the IEEE/CVF International Conference on
  Computer Vision}, pages 3713--3722, 2019.

\bibitem{zhang2018deep}
Ying Zhang, Tao Xiang, Timothy~M Hospedales, and Huchuan Lu.
\newblock Deep mutual learning.
\newblock In {\em Proceedings of the IEEE conference on computer vision and
  pattern recognition}, pages 4320--4328, 2018.

\bibitem{zhao2022decoupled}
Borui Zhao, Quan Cui, Renjie Song, Yiyu Qiu, and Jiajun Liang.
\newblock Decoupled knowledge distillation.
\newblock In {\em Proceedings of the IEEE/CVF Conference on Computer Vision and
  Pattern Recognition}, pages 11953--11962, 2022.

\bibitem{zhao2021similarity}
Haoran Zhao, Kun Gong, Xin Sun, Junyu Dong, and Hui Yu.
\newblock Similarity transfer for knowledge distillation.
\newblock {\em arXiv preprint arXiv:2103.10047}, 2021.

\bibitem{9711488}
Yichen Zhu and Yi Wang.
\newblock Student customized knowledge distillation: Bridging the gap between
  student and teacher.
\newblock In {\em 2021 IEEE/CVF International Conference on Computer Vision
  (ICCV)}, pages 5037--5046, 2021.

\end{thebibliography}
\small
\bibliographystyle{./ieee_fullname}

\end{document}